\documentclass[sigconf, nonacm]{acmart}

\usepackage{multirow}

\begin{document}

\newcommand*{\fname}{DataSculpt}

\title{Can Large Language Models Design Accurate Label Functions?}

\author{Naiqing Guan}
\affiliation{%
  \institution{University of Toronto}
  \city{Toronto}
  \country{Canada}
}
\email{naiqing.guan@mail.utoronto.ca}

\author{Kaiwen Chen}
\affiliation{%
  \institution{University of Toronto}
  \city{Toronto}
  \country{Canada}
}
\email{kckevin.chen@mail.utoronto.ca}

\author{Nick Koudas}
\affiliation{%
  \institution{University of Toronto}
  \city{Toronto}
  \country{Canada}
}
\email{koudas@cs.toronto.edu}

\begin{abstract}
Programmatic weak supervision methodologies facilitate the expedited labeling of extensive datasets through the use of label functions (LFs) that encapsulate heuristic data sources. Nonetheless, the creation of precise LFs necessitates domain expertise and substantial endeavors. Recent advances in pre-trained language models (PLMs) have exhibited substantial potential across diverse tasks. However, the capacity of PLMs to autonomously formulate accurate LFs remains an underexplored domain. In this research, we address this gap by introducing \fname{}, an interactive framework that harnesses PLMs for the automated generation of LFs. Within \fname{}, we incorporate an array of prompting techniques, instance selection strategies, and LF filtration methods to explore the expansive design landscape. Ultimately, we conduct a thorough assessment of \fname{}'s performance on 12 real-world datasets, encompassing a range of tasks. This evaluation unveils both the strengths and limitations of contemporary PLMs in LF design.
\end{abstract}

\maketitle

\section{Introduction}
Modern machine learning models usually require large amounts of training data to achieve good accuracy. However, manually labelling data is costly, especially when domain expertise is required. Recently proposed weak supervision approaches, such as data programming \cite{ratner2016data,ratner2017snorkel}, provide a promising alternative for labelling large datasets rapidly. In the data programming paradigm, instead of manually labelling instances one by one, users design label functions (LFs) that can provide noisy labels to a subset of data at a much cheaper cost. Then a label model learns the accuracies of the LFs to denoise and aggregate the weak labels. Finally, the labels predicted by the label model are used to train the downstream machine learning model. The data programming paradigm has been successfully applied to various domains \cite{bach2019snorkel}.

Although data programming saves the bulk of the labelling effort, it still requires nontrivial effort to design accurate label functions. Typically one is required to select a small subset of unlabeled data as the development data, identify patterns from these data, and represent the patterns in the form of LFs. In order to save the effort required in the LF development process, researchers have designed frameworks to automate the process, such as Snuba \cite{varma2018snuba}, or guiding users to design LFs, such as IWS\cite{boecking2020interactive} and Nemo \cite{hsieh2022nemo}. However, Snuba only supports simple heuristics like decision stumps and falls short in capturing rich semantics in the text,  while IWS and Nemo require interaction with human experts in the loop. Therefore, how to automatically generate accurate LFs for data programming still requires further investigation.

The recently emerging pre-trained language models (PLMs) like GPT-3 \cite{brown2020language} and GPT-4 \cite{openai2023gpt4} have been shown to achieve strong performance in various tasks, such as translation, question answering, and cloze tasks. There are also researches demonstrating their potential in data wrangling tasks \cite{narayan2022can,jaimovitch2023can}. The successes of PLMs in various tasks indicate that they may also help automate the LF design process and alleviate human burden. However, to the best of our knowledge, there is currently no research work that systematically investigates this opportunity. Naturally, one may wonder why we still need data programming for labelling datasets, given that we can prompt the PLMs to label data directly.  First, as many PLMs \cite{brown2020language,openai2023gpt4} charge by tokens, it would be costly to annotate large datasets by querying every instance with prompting. Data programming is a cost effective alternative for labelling large datasets, provided it performs well, compared to alternatives. Secondly, data programming leverages label functions to annotate datasets, which are more interpretable compared to the labels provided by PLMs. In this work, we bridge the gap mentioned above and explore the opportunity of leveraging PLMs to design LFs automatically. We aim to answer the following research questions:

\begin{description}
    \item [RQ1:] In which cases can large language models design accurate label functions? How is the quality of generated label functions compared to human-designed ones?

    \item [RQ2:] Various prompt methods such as chain-of-thought \cite{wei2022chain}, self-consistency \cite{wang2022self}, and in-context example selection \cite{liu2021makes} have been proposed to improve PLM's performance. Do these methods benefit LF development?

    \item [RQ3:] Which query instances should we select in the prompts to design LFs? How will different selection methods affect the end-to-end system performance?

    \item [RQ4:] 
    What considerations should be taken into account when employing PLMs to prompt the automatic design of LFs?

\end{description}

To explore these research questions, we built an interactive data programming framework \fname{} \footnote{Code available at \url{https://github.com/Gnaiqing/LLMDP}}, which prompts PLMs to automatically generate LFs. We extensively evaluated \fname{} in 12 real-world datasets, covering various tasks such as spam detection, sentiment analysis, topic classification and relation classification. We experimented with various prompting methods, pre-trained language models, query instance selection methods, and LF filtering methods in \fname{} to explore the design space. Finally, we provide a comprehensive experiment analysis to identify the strengths and limitations of PLMs in designing LFs. 

To summarize, our contributions are as follows:
\begin{itemize}
    \item We designed an interactive data programming framework \fname{} for automatic LF development via PLM prompting.

    \item We designed and implemented various strategies for the modules in \fname{}, such as prompt design, query instance selection and LF filtering, to support a comprehensive evaluation.

    \item We evaluated \fname{} on 12 real-world datasets to measure the quality of generated LFs and the end-to-end framework performance. We also conducted comparative studies to explore the design space from various aspects.

    \item We provided insights on when PLMs can design accurate LFs and when they fall short. We also highlighted important and unresolved research issues in leveraging PLMs for automating the LF development process.
\end{itemize}

\section{background}
\subsection{Data Programming}
Data programming (DP) \cite{ratner2016data,ratner2017snorkel} is an emerging framework that leverages various weak supervision sources to label training data automatically. In the DP framework, users first design label functions to provide weak labels to the data, and then the label model aggregates noisy, weak labels to provide probabilistic labels to the unlabeled data. Researchers have proposed various label models \cite{ratner2017snorkel,ratner2019training,fu2020fast} to learn the accuracy and correlations between label functions. We refer interested readers to a recent survey \cite{zhang2022survey} for a comprehensive discussion on label models and data programming in general. 

Label function design is an important yet understudied area, partly because designing label functions involves extensive human efforts. Snuba \cite{varma2018snuba} proposes automatically generating label functions by learning simple models over a small labelled dataset. IWS \cite{boecking2020interactive}, WITAN \cite{denham2022witan} and Darwin \cite{galhotra2021adaptive} interactively ask users to verify candidate label functions. Nemo \cite{hsieh2022nemo} interactively selects candidate instances for users to observe and develop label functions. 

\subsection{Pre-trained Language Models}
Pre-trained language models (PLMs), such as BERT \cite{devlin2018bert}, RoBERTa \cite{liu2019roberta} and XLNet \cite{yang2019xlnet}, have been successfully applied in various NLP tasks and made impressive progress. 
More recent PLMs, such as the GPT series \cite{brown2020language} developed by OpenAI and the Llama series developed \cite{touvron2023llama} by meta, can be applied to many downstream tasks directly without fine-tuning. 
To utilize such models, one can provide prompts with a few examples. This few-shot prompting has attracted great attention and proven to be effective on many tasks. As the prompt design has a great impact on the PLM's performance over specific tasks, researchers have investigated various prompt engineering methods \cite{wei2022chain,wang2022self,liu2021makes} to improve the model's performance. Our work evaluated several well-known prompt engineering methods in the context of LF development.

\section{Framework Description}\label{sec:framework}

\begin{figure}[htbp]
\centering
 \includegraphics[width=\columnwidth]{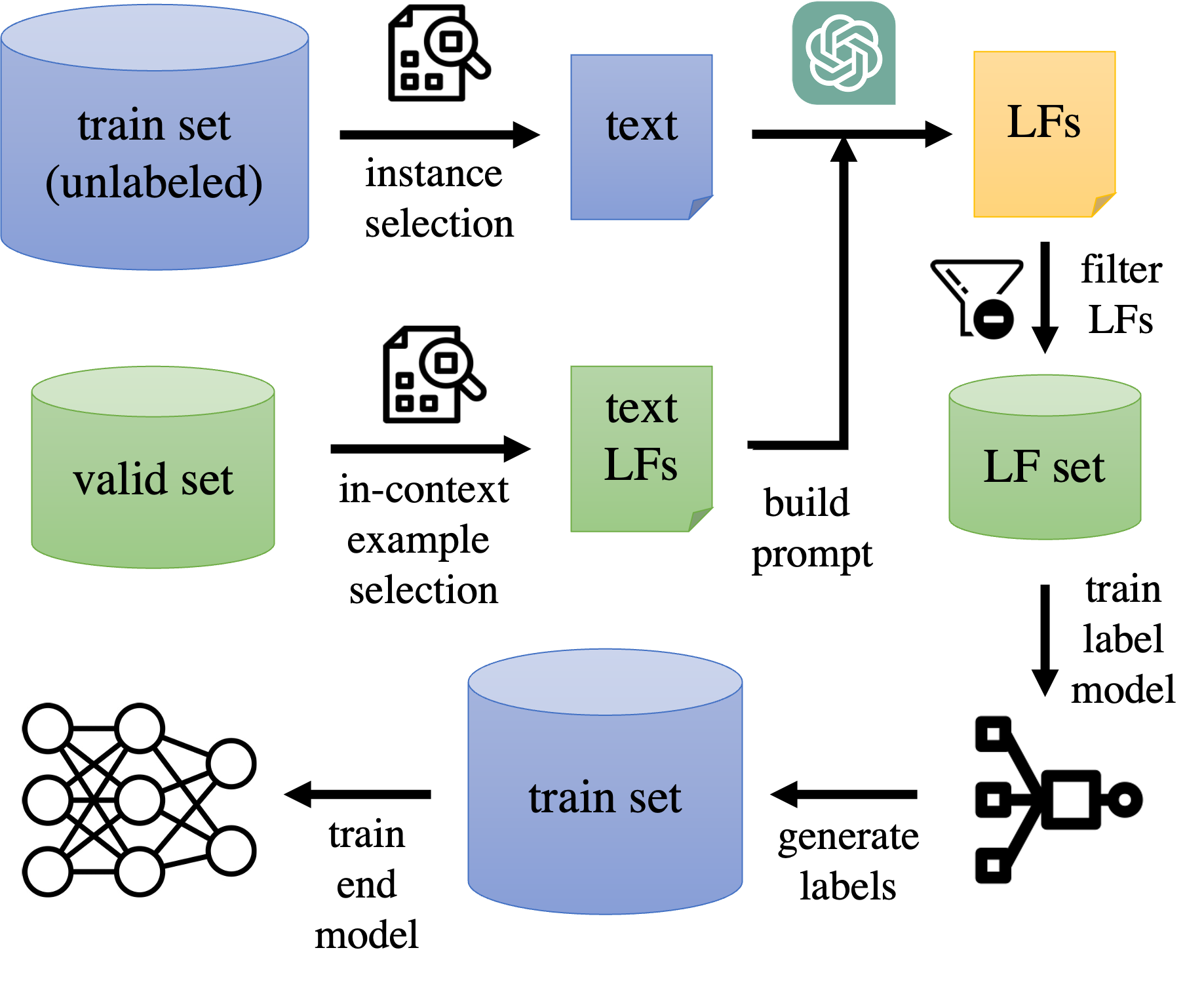}
 \caption{Workflow of the \fname{} framework.}\label{fig:workflow}
\end{figure}

Figure \ref{fig:workflow} illustrates the workflow of \fname{}. The framework creates LFs interactively. In each iteration, \fname{} first selects an instance from the unlabeled data. Then, it builds a prompt asking the PLM to provide LFs based on the instance.  After the LFs are extracted from PLM responses, they are verified based on criteria such as validity, accuracy and redundancy. The verified LFs are then added to the LF set achieved in previous iterations. Then the label model aggregates the LF outputs together, and the downstream model is trained using the labels generated by the label model. 

\subsection{LF space}

\fname{} supports two kinds of tasks: text classification and relation classification. The text classification task aims to classify a certain passage into multiple groups, such as topic classification or sentiment classification. The relation classification tasks, on the other hand, aim at classifying the relationship between two entities given a passage, such as deciding whether there is a causal relationship between two events. 

The label functions can come from various sources, such as manually designed heuristics, crowdsourced labels or distant supervision by existing knowledge bases \cite{ratner2017snorkel}. In \fname{}, we focus on two kinds of label functions commonly used in previous works \cite{yu2020fine,hsieh2022nemo}: keyword-based LFs and pattern-based LFs. A keyword-based LF $\lambda_{k,c}$ labels the given passage as class $c$ when the passage contains a specific keyword (or phrase) $k$. A pattern-based LF $\lambda_{p,c}$ labels the given passage as class $c$ when a specific text pattern $p$ matches the passage. In \fname{}, we restrict the keywords to be unigram to trigram phrases and the text pattern to be ones representable using regular expressions. 

The keyword-based LFs usually work well for text classification, but they are insufficient for relation classification. For example, a keyword-based LF with keyword \textit{cause} cannot distinguish between \textit{A cause B} and \textit{A cause C}. Therefore, \fname{} generates keyword-based LFs for text classification tasks and pattern-based LFs for relation classification tasks.

\subsection{Prompt Template}

\begin{figure*}[t]
\centering
 \includegraphics[width=2\columnwidth]{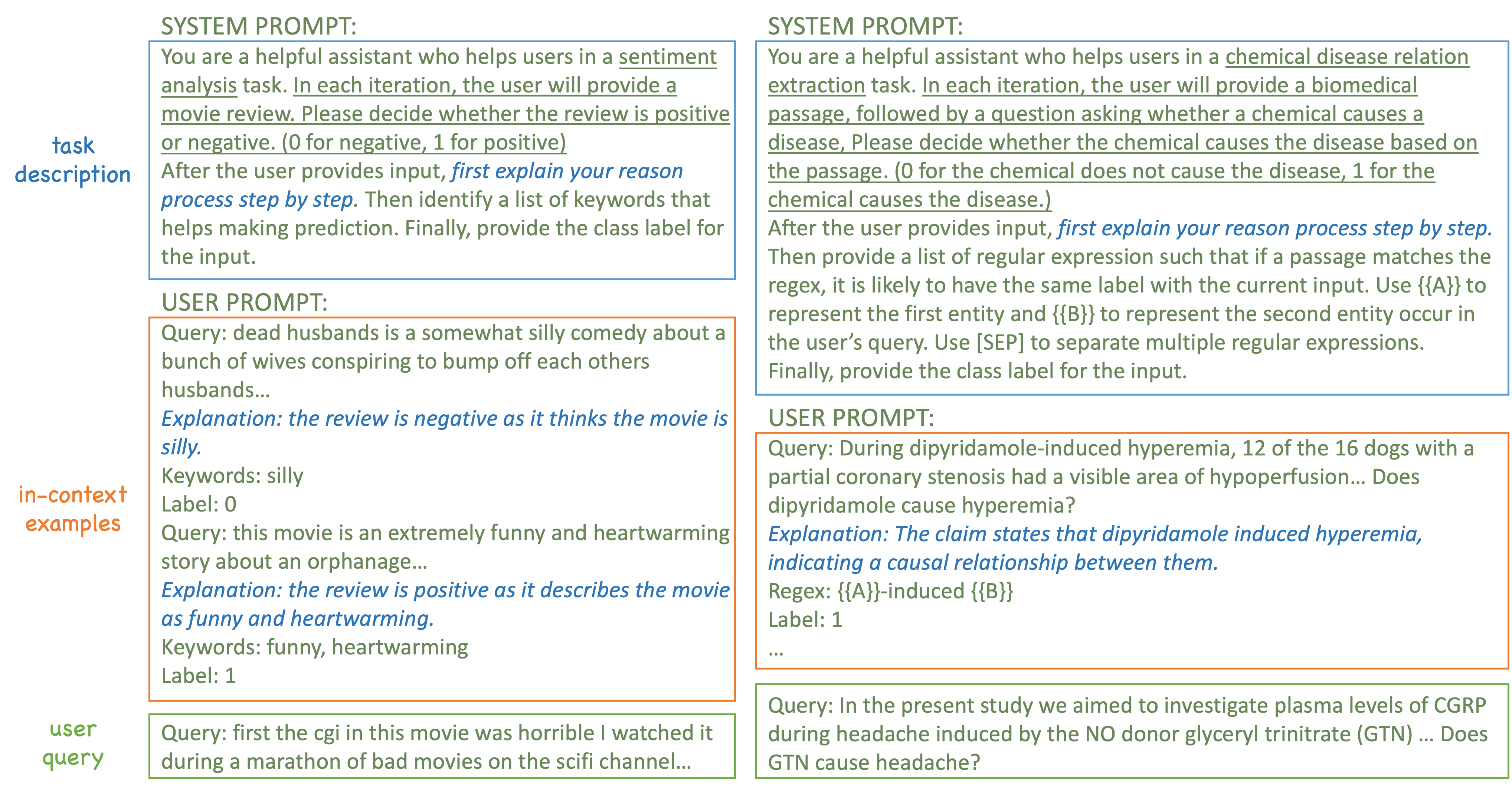}
 \caption{Prompt templates for the \fname{} framework. The underlined instructions are dataset-specific. The italic instructions are optional for chain-of-thought prompting.}\label{fig:prompt}
\end{figure*}

\fname{} supports two different prompt templates for keyword-based LFs and pattern-based LFs, respectively. Figure \ref{fig:prompt} illustrates the templates. Each template starts with a task description with the definition of separate classes. Then it provides an interactive format for PLMs to follow. For the keyword-based prompt template, we provide a paragraph, and then the PLM responds with the predicted label and a set of keywords that helps make classification. For the pattern-based prompt template, we provide a passage, followed by a question asking about the relationship between two entities that occurred in the passage. Then the PLM provides the predicted label and a set of regular expressions that are indicative of the predicted label. When no indicative keywords or regular expressions can be identified, the PLM returns the predictive label only.

Previous work \cite{wei2022chain} demonstrates that requesting the PLM to provide a chain of thought (CoT) before making a final prediction helps improve response accuracy. To evaluate the effect of the chain of thought in generating label functions, we optionally ask the PLM to provide a step-by-step reasoning process before predicting the label. The lines in italic in Figure \ref{fig:prompt} are included when CoT is applied. 

\subsection{In-context Example Selection}
In the user prompt, providing a few in-context examples to construct the context before the query can help the PLM better understand the task and improve its effectiveness. Previous research has shown that the performance of PLM is sensitive to the choice of in-context examples \cite{brown2020language,liu2021makes}. In \fname{}, we evaluate two in-context example selection methods: \textit{class-balanced}, which randomly selects in-context examples in a class-balanced way (k examples per class), and \textit{KATE} \cite{liu2021makes}, which selects the examples that are mostly close to the test input in a feature space. We select the in-context examples from a labelled validation set drawn i.i.d. from the unlabeled training set. 

Note that apart from the label, the examples need to provide a set of indicative keywords or regular expressions and a step-by-step reasoning process (if applying CoT). However, this information needs manual annotation. For class-balanced example selection, we select a fixed set of examples from the validation set and manually provide the keywords, regular expressions and explanations. For KATE \cite{liu2021makes}, as different query instances require different in-context examples, manual annotation would be infeasible. Inspired by previous works on automatically generating chain-of-thought \cite{kojima2022large,zhang2022automatic}, we apply the PLM to automatically provide indicative keywords, regular expressions and chain-of-thought for the in-context examples selected by the KATE sampler. As the in-context examples have labels provided in advance, we ask the PLM to provide a step-by-step reasoning process and identify indicative keywords or patterns to justify the provided label. The prompt we use is similar to the prompts in Figure \ref{fig:prompt}, except that the labels are provided in the user input instead of PLM responses. 

\subsection{Query Instance Selection}
\fname{} iteratively selects a subset of instances for PLMs to check and develop LFs, and the selected instances can have a great influence on the quality of LFs. In \fname{}, we evaluate several instance selection methods:
\begin{itemize}
\item Random Sampling: randomly select instances from the unlabeled dataset to develop LFs. This is the default strategy applied in \fname{}.

\item Uncertain Sampling \cite{lewis1995sequential}: select the instances where the current downstream model is most uncertain. The uncertainty is measured using predictive entropy.

\item Select by Expected Utility (SEU) \cite{hsieh2022nemo}: selects the instance by:
\begin{equation*}
x^*=argmax_{x\in U} E_{P(\lambda|x)}[\Psi_t(\lambda)]
\end{equation*}
where $P(\lambda|x)$ is the probability that a user will return the LF $\lambda$ when observing instance $x$, and $\Psi_t(\lambda)$ is the utility of the LF $\lambda$ to the current LF set. SEU builds different models to estimate the probability and the utility. We refer readers to the original paper \cite{hsieh2022nemo} for more details.

\end{itemize}

\subsection{LF Filtering}
As the PLM can make mistakes or generate invalid outputs occasionally, an LF filtering step is required to ensure the quality of generated LFs. We apply several filters in the LF filtering step:

\begin{itemize}
\item Validity Filter: the validity filter checks the validity of candidate label functions, such as the validity of regular expressions generated by the label model. \fname{} checks that the keyword is either unigrams, bigrams or trigrams, the regular expression is valid, and the label is in the candidate classes.

\item Accuracy Filter: the accuracy filter evaluates the accuracy of candidate label functions on the validation set and prunes out label functions with accuracy below a certain threshold. In our experiments, we set the accuracy threshold as 0.6 by default. When the candidate LF is not active on any instance of the validation set, we assume the LF passes the accuracy filter.

\item Redundancy Filter: the redundancy filter prunes out the candidate label functions whose output has high similarity with other label functions in the LF set. We use an intersection over union metric for the redundancy filter, which prunes out an LF if its consensus with an existing LF is above 95\% over the instances where either LF is active.
\end{itemize}

\subsection{Default Class} 
In some classification tasks, the positive class refers to the existence of some pattern, and the negative class refers to the absence of it. In these tasks, it is usually much easier to design LFs for the positive class than the negative class. One example is the chemical-disease relationship classification task in Figure \ref{fig:prompt}. While it is easy to identify patterns that indicate that the chemical causes the disease, it is difficult to identify patterns that indicate the absence of such a relationship. In our evaluation, we find that PLMs usually refuse to provide LFs for the negative class in that case, which deteriorates the performance of data programming due to class imbalance in the LF set. To deal with this problem, we define a \textit{default class} for datasets with the above characteristics. When a default class is defined, if an instance is not covered by any LFs, it will be assigned to the default class before training the downstream model. In our evaluated datasets, we define the default class for the spouse dataset (default is no spouse relationship) and the CDR dataset (default is no causal relationship).

\section{Experiments}
In this section, we provide a comprehensive evaluation of \fname{} to investigate the performance of large language models in generating label functions. 

\begin{table}[t]
\footnotesize
\centering
\caption{Datasets used in Evaluation.} \label{tab:datasets}
\vspace{-0.1cm}
\resizebox{\columnwidth}{!}{
\begin{tabular}{c c c c c c c}
\toprule[1.5pt]
\textbf{Task} & \textbf{Domain} & \textbf{Dataset} & \textbf{\#Class} & \textbf{\#Train} & \textbf{\#Valid} & \textbf{\#Test} \\
\hline
\multirow{2}{*}{Spam Cls.} & Review & Youtube \cite{alberto2015tubespam} & 2 & 1586 & 120 & 250 \\
\cline{2-7}
& Text Message & SMS \cite{almeida2011contributions,awasthi2019learning} & 2 & 4571 & 500 & 500 \\
\hline
\multirow{2}{*}{Sentiment Cls.} & Movie & IMDB \cite{maas2011learning,ren2020denoising} & 2 & 20000 & 2500 & 2500 \\
\cline{2-7}
& Review & Yelp \cite{zhang2015character,ren2020denoising} & 2 & 30400 & 3800 & 3800 \\
\hline
\multirow{3}{*}{Topic Cls.} & News & Agnews \cite{zhang2015character,ren2020denoising} & 4 & 96000 & 12000 & 12000 \\
\cline{2-7}
& Paper Abstract & ArxivAbs \cite{arxivabstract} & 2 & 21367 & 4579 & 4579 \\
\cline{2-7}
& Biomedical & MedAbs \cite{schopf2022evaluating} & 5 & 8085 & 3465 & 2888\\
\hline
Question cls. & Web Query & TREC \cite{li2002learning,awasthi2019learning} & 6 & 4965 & 500 & 500 \\
\hline
\multirow{4}{*}{Relation Cls.} & News & Spouse \cite{corney2016million,ratner2017snorkel} & 2 & 22254 & 2811 & 2701 \\
\cline{2-7}
& Biomedical & CDR \cite{davis2017comparative,ratner2017snorkel} & 2 & 8430 & 920 & 4673 \\
\cline{2-7}
& Web Text & SemEval \cite{hendrickx2010semeval,zhou2020nero} & 9 & 1749 & 200 & 692 \\
\cline{2-7}
& Chemical & ChemProt \cite{krallinger2017overview,yu2020fine} & 10 & 12861 & 1607 & 1607 \\
\bottomrule[1.5pt]
\end{tabular}}
\end{table}

\begin{table*}[t]
\footnotesize
\centering
\caption{LF statistics and model performance with different prompting methods.} \label{tab:eval}
\vspace{-0.1cm}
\resizebox{2\columnwidth}{!}{
\begin{tabular}{c c c c c c c c c c c c c c}
\toprule[1.5pt]
\textbf{Metric} & \textbf{Method} & \textbf{Youtube} & \textbf{SMS(F1)} & \textbf{IMDB} & \textbf{Yelp} & \textbf{Agnews} & \textbf{TREC} & \textbf{ArxivAbs(F1)} & \textbf{MedAbs} & \textbf{ChemProt} & \textbf{CDR(F1)} & \textbf{Spouse(F1)} & \textbf{SemEval} \\
\hline
\multirow{4}{*}{PLM\_acc} & Few-Shot & 80.00 (5.83) & \textbf{97.60 (3.28)} & 89.60 (0.89) & \textbf{96.80} (1.78) & \textbf{91.20 (7.56)} & 71.60 (5.17) & 89.60 (4.56) & 63.6 (9.09) & 28.4 (3.28) & 66.4 (7.66) & -- & 66.00 (11.31) \\
& CoT & 82.40 (7.40)& 86.00 (3.16) & 91.60 (3.57) & 95.60 (1.67) & 82.80 (5.21) & \textbf{81.60 (2.96)} & 88.80 (3.89) & 60.40 (10.62) & 41.60 (7.40) & 65.60 (7.79) & -- & 74.80 (6.09) \\
& SC &  83.20 (7.82) & 91.20 (3.03) & 93.20 (1.78) & \textbf{96.80 (1.09)} & 86.80 (5.40) & 78.80 (6.41) & 90.00 (3.74) & \textbf{65.60 (9.20)} & 40.00 (9.48) & 63.20 (7.82) & -- & 80.40 (5.89) \\
& SC+KATE & \textbf{84.00 (6.00)} & 91.20 (3.63) & \textbf{94.40 (2.60)} & \textbf{96.80 (2.28)} & 85.20 (7.56) & 77.20 (3.89) & \textbf{92.80 (2.28)} & 58.80 (8.67) & \textbf{61.20 (5.76)} & \textbf{71.60 (9.94)} & -- & \textbf{80.80 (5.21)} \\
\hline
\multirow{5}{*}{LF\_num} & Wrench & 10 & 73 & 5 & 8 & 9 & 68 & -- & -- & 26 & 33 & 9 & 164 \\
& Few-Shot & 75.60 (17.98) & 163.80 (26.66) & 87.60 (17.32) & 127.40 (6.09) & 180.40 (19.11) & 19.40 (2.60) & 91.80 (13.53) & 56.80 (6.79) & 16.20 (4.43) & 17.20 (10.75) & 19.40 (3.50) & 11.40 (3.20) \\
& CoT &  68.80 (6.14) & 115.20 (4.14) & 103.80 (20.77) & 117.60 (13.42) & 158.20 (15.17) & 31.60 (3.50) & 90.20 (11.25) & 52.60 (5.81) & 20.60 (4.33) & 7.40 (2.51) & 6.00 (4.06) & 14.60 (2.88) \\
& SC &  107.80 (9.57) & \textbf{201.00 (23.70)} & 225.40 (45.60) & 246.60 (39.40) & 225.00 (20.40) & 73.60 (3.78) & 178.00 (19.82) & 87.40 (4.27) & \textbf{50.80 (8.22)}  & 28.80 (9.75) & \textbf{23.80 (17.41)} & \textbf{30.40 (4.15)} \\
& SC+KATE & \textbf{116.80 (8.25)} & 199.80 (31.30) & \textbf{329.40 (28.15)} & \textbf{320.80 (22.27)} & \textbf{236.00 (19.19)} & \textbf{94.80 (10.40)} & \textbf{209.00 (24.54)} & \textbf{100.60 (9.07)} & 48.60 (8.20) & \textbf{33.80 (6.09)} & 7.70 (1.70) & 27.40 (5.32) \\
\hline
\multirow{5}{*}{LF\_acc\_avg} & Wrench & 83.16 & 97.26 & 69.88 & 73.05 & 81.66 & 75.92 & -- & -- & 46.65 & 75.27 & -- & 97.69 \\
& Few-Shot & 73.91 (9.79) & \textbf{91.34 (1.96)} & 71.75 (2.08) & 76.93 (2.25) & \textbf{84.78 (3.08)} & \textbf{82.47 (10.05)} & \textbf{88.77 (1.04)} & \textbf{62.21 (3.13)} & 46.08 (13.46) & \textbf{55.9 (2.89)} & -- & \textbf{72.53 (11.38)} \\
& CoT & \textbf{75.65 (2.99)} & 86.13 (1.02) & 72.81 (1.75) & \textbf{77.54 (1.16)} & 82.59 (4.42) & 81.29 (9.77) & 88.74 (1.01) & 61.85 (6.37) & 51.77 (6.43) & 46.03 (17.32) & -- & 69.95 (11.78) \\
& SC &  73.50 (4.40) & 88.36 (1.03) & 72.61 (1.76) & 77.18 (1.30) & 82.41 (3.82) & 74.25 (5.73) & 87.27 (2.05) & 61.73 (5.22) & 49.53 (3.88) & 45.26 (6.80) & -- & 71.83 (6.99) \\
& SC+KATE & 72.63 (5.33) & 86.99 (2.49) & \textbf{73.16 (0.95)} & 76.06 (1.90) & 81.09 (6.04) & 68.26 (6.15) & 87.89 (1.22) & 57.75 (6.81) & \textbf{53.08 (6.34)} & 53.05 (6.04) & --  & 60.72 (5.74) \\
\hline
\multirow{5}{*}{LF\_cov\_avg} & Wrench & 16.34 & 0.71 & 23.60 & 18.34 & 10.34 & 2.54 & -- & -- & 5.93 & 6.26 & 3.75 & 0.76 \\
& Few-Shot & 2.46 (0.41) & 0.73 (0.02) & \textbf{3.76 (0.64)} & \textbf{2.35 (0.26)} & 0.23 (0.04) & 0.69 (0.09) & \textbf{3.92 (0.77)} & \textbf{0.62 (0.09)} & 0.99 (0.34) & 0.98 (1.30) & 2.78 (1.48) & \textbf{7.34 (1.64)} \\
& CoT & \textbf{2.64 (0.35)} & 0.74 (0.12) & 3.37 (0.32) & 2.25 (0.31) & 0.30 (0.05) & \textbf{0.76 (0.15)} & 3.91 (0.50) & 0.47 (0.20) & \textbf{1.54 (0.44)} & 0.26 (0.19) & \textbf{3.82 (5.23)} & 4.76 (0.82) \\
& SC & 2.14 (0.37) & \textbf{0.86 (0.14)} & 2.35 (0.40) & 1.75 (0.25) & \textbf{0.32 (0.03)} & 0.49 (0.07) & 3.47 (0.67) & \textbf{0.62 (0.16)} & 1.01 (0.10) & 1.21 (1.81) & 1.16 (1.12) & 4.57 (0.63) \\
& SC+KATE & 1.96 (0.16) & 0.76 (0.13) & 1.71 (0.16) & 1.40 (0.14) & 0.27 (0.02) & 0.39 (0.04) & 3.37 (0.39) & 0.38 (0.06) & 1.06 (0.08) & \textbf{1.51 (1.71)} & 0.50 (0.52) & 4.28 (1.28) \\
\hline
\multirow{5}{*}{Train\_acc} & Wrench & 81.83 (4.74) & 93.75 (4.48) & 74.63 (1.00) & 74.11 (0.03) & 81.91 (0.28) & 50.62 (1.00) & -- & -- & 57.38 (0.04) & 73.09 (0.56) & -- & 74.61 (0.38) \\
& Few-Shot &  \textbf{87.34 (2.20)} & 92.45 (1.00) & 78.27 (0.70) & 79.21 (3.40) & \textbf{82.17 (2.74)} & \textbf{96.34 (1.52)} & \textbf{77.95 (5.83)} & 59.31 (19.53) & 55.95 (27.03) & 62.81 (0.46) & -- & 81.26 (3.79) \\
& CoT & 85.27 (2.13) & 91.49 (1.65) & 78.60 (1.08) & 76.35 (2.91) & 80.13 (2.55) & 94.74 (2.30) & 69.80 (22.86) &  \textbf{68.43 (1.50)} & 71.93 (9.06) & 62.37 (0.22) & -- & 79.59 (1.01) \\
& SC & 82.64 (3.05) & \textbf{92.50 (0.44)} & \textbf{79.89 (0.77)} & 81.45 (1.91) & 75.93 (9.77) & 82.65 (17.66) & 73.18 (15.78) & 58.62 (20.68) & 70.27 (9.53) & 62.79 (0.80) & -- & \textbf{83.04 (1.55)} \\
& SC+KATE &  82.16 (1.43) & 90.75 (2.22) & 79.01 (1.12) & \textbf{81.58 (1.42)} & 79.24 (1.96) & 85.98 (2.71) & 62.15 (18.88) & 56.89 (20.86) & \textbf{74.43 (0.84)} & \textbf{63.72 (2.11)} & -- & 78.32 (3.78) \\
\hline
\multirow{5}{*}{Train\_cov} & Wrench & 87.70 & 40.52 & 87.58 & 82.78 & 69.08 & 95.13 & -- & -- & 85.62 & 90.72 & 25.77 & 100.00 \\
& Few-Shot & 80.76 (2.29) & 63.73 (6.20) & 93.17 (2.14) & 89.99 (2.11) & 32.25 (5.10) & 13.49 (3.36) & 96.84 (2.14) & 26.46 (3.44) & 15.02 (8.58) & 100.0 (0.00) & 100.0(0.00) & 57.71 (14.56) \\
& CoT & 78.17 (2.17) & 52.86 (6.78) & 94.23 (1.48) & 87.12 (3.27) & 34.98 (6.03) &22.27 (3.24) & 97.19 (0.91) & 20.67 (9.56) & 26.77 (4.07) & 100.0(0.00) & 100.0 (0.00) & 45.64 (15.00) \\
& SC &  81.97 (1.48) & \textbf{75.38 (6.03)} & 97.73 (0.60) & 94.65 (1.58) & \textbf{47.99 (4.42)} & 29.84 (2.52) & 99.69 (0.20) & \textbf{36.45 (5.74)} & 31.00 (3.08) & 100.0 (0.00) & 100.0 (0.00) & \textbf{77.78 (7.28)} \\
& SC+KATE & \textbf{83.18 (1.49)} & 71.58 (7.67) & \textbf{98.04 (0.53)} & \textbf{94.78 (1.06)} & 43.82 (4.40) & \textbf{30.80 (3.35)} & \textbf{99.76 (0.17)} & 27.73 (5.74) & \textbf{35.42 (1.75)} & 100.0 (0.00) & 100.0 (0.00) & 63.68 (6.71) \\
\hline
\multirow{5}{*}{Test\_acc/F1} & Wrench & 88.72 (0.65) & 86.46 (2.05) & 76.13 (0.63) & 84.73 (0.27) & 83.53 (0.31) & 58.92 (1.98) & -- & -- & 54.29 (0.09) & 54.69 (0.79) & 18.10 (0.00) & 72.57 (0.14) \\
& Few-Shot & \textbf{89.60 (1.96)} & 82.11 (2.02) & 79.14 (0.44) & 86.86 (0.88) & \textbf{81.25 (1.13)} & 44.48 (6.86) & 56.73 (18.85) & 37.50 (2.87) & 27.32 (14.7) & 15.75 (10.18) & 34.51 (1.37) & 60.80 (12.58) \\
& CoT & 88.48 (2.95) & 81.72 (6.46) & 79.33 (0.87) & 85.39 (1.24) & 79.48 (2.65) & 47.60 (6.85) & 55.43 (19.31) & 37.56 (0.83) & 39.44 (5.18) & 15.24 (8.10) & 24.56 (19.24) & 49.57 (14.43) \\
& SC & 87.92 (3.48) & 82.88 (2.89) & \textbf{79.39 (0.41)} & 87.39 (0.67) & 75.84 (11.82) & 45.96 (8.57) & \textbf{57.95 (15.59)} & \textbf{39.72 (0.85)} & 41.84 (6.44) & \textbf{21.65 (11.61)} & \textbf{36.25 (9.08)} & \textbf{81.10 (5.27)} \\
& SC+KATE & 86.24 (6.25) & \textbf{83.65 (3.62)} & 78.81 (0.75) & \textbf{87.87 (0.39)} & 79.98 (2.41) & \textbf{50.52 (7.06)} & 45.23 (16.95) & 37.67 (3.06) & \textbf{51.33 (2.90)} & 19.91 (16.06) & 32.36 (24.23) & 64.43 (5.63) \\
\bottomrule[1.5pt]
\end{tabular}}
\end{table*}

\subsection{Experiment Setup}
We evaluate \fname{} on 12 real-world text datasets, 10 of which come from the WRENCH benchmark \cite{zhang2021wrench}, covering tasks including spam detection, sentiment analysis, topic classification and relation classification. We omit the label functions associated with these datasets in the WRENCH benchmark and automatically create label functions using \fname{}. Apart from these datasets, we additionally include two datasets to better understand the performance of \fname{} in domain-specific settings: MedAbs \cite{schopf2022evaluating}, which classify the patient's disease into one of five categories based on a prescription, and ArxivAbs \cite{arxivabstract}, which classify an arxiv paper into the domain of computer vision or machine learning based on its title and abstract. Table \ref{tab:datasets} provides a summary of the datasets used in our evaluation.

\begin{table}[t]
\footnotesize
\centering
\caption{Metrics used in evaluation.} \label{tab:metrics}
\vspace{-0.1cm}
\resizebox{\columnwidth}{!}{
\begin{tabular}{c c}
\toprule[1.5pt]
\textbf{Metric} & \textbf{Description}\\
\hline
PLM\_acc & Accuracy of labels provided by PLM in the responses  \\
\hline
LF\_num & Number of generated LFs in the final LF set \\
\hline
LF\_acc\_avg & Average LF accuracy in the final LF set \\
\hline
LF\_cov\_avg & Average LF coverage in the final LF set \\
\hline
Train\_acc & The label model's accuracy on the training data (on the part covered by LFs) \\
\hline
Train\_cov & The percentage of training data covered by LFs \\
\hline
Test\_acc/F1 & The downstream model's accuracy (F1 score for imbalanced datasets) on the test data \\
\bottomrule[1.5pt]
\end{tabular}}
\end{table}

For each run of \fname{}, we iteratively select 50 query instances to design label functions.  We do not restrict the number of LFs returned in each iteration, so the total number of generated LFs can be greater than 50. We repeat each run 5 times using different random seeds and report the average LF statistics and downstream model performance after the LF development process. We also report standard deviations in the parenthesis. We use BERT \cite{devlin2018bert} to extract features from the text, MeTaL \cite{ratner2019training} as the label model and logistic regression as the downstream model, which are consistent with the configurations in the WRENCH benchmark \cite{zhang2021wrench}. To evaluate \fname{} comprehensively, we use various metrics to measure the quality of LFs, training labels generated by the label model, and the performance of the downstream model.  Note that the ground truth labels for the training data are not available for the Spouse dataset, so we only report metrics that do not require ground truth training labels on that dataset. Table \ref{tab:metrics} describes the metrics we use in evaluation. 

\fname{} has various configuration options that affect its performance. By default, we use the \textit{gpt-3.5-turbo-0613} model supported by the OpenAI API \cite{openai2023} as the PLM to query, apply random sampling for query instance selection, class-balanced sampling with 1 example per class for in-context example selection, and apply all three filters (validity, accuracy, redundancy) in the LF filtering step. We will evaluate the effect of each component in subsequent sections. 

Apart from LFs developed by \fname{}, we also evaluated the LFs provided in the WRENCH benchmark \cite{zhang2021wrench}, which are designed by human experts. The evaluation is conducted using the same label model and downstream model as \fname{}. Note that the LFs provided in WRENCH were designed by different human experts under different LF development scenarios. Therefore, the evaluation results on WRENCH may not reflect the quality of manually designed LFs if different experts are employed, or the devoted efforts change. Nonetheless, the addition of this baseline provides an intuitive comparison of the quality of automatically generated LFs compared to manually designed ones in a well-accepted benchmark.

\subsection{Performance Overview}
To investigate in which cases can PLMs design accurate LFs, we first evaluate the basic few-shot prompt version of \fname{} without chain-of-thought or other advanced prompting techniques. The performance is illustrated in the \textit{Few-Shot} rows in Table \ref{tab:eval}. We first look at the \textit{LF\_acc\_avg} metric, which evaluates the average accuracy of generated LFs. For tasks that require general knowledge, such as sentiment analysis and spam detection, the accuracy of generated LFs is usually high (> 70\%). However, for tasks that require specific expertise, such as biomedical knowledge, the accuracy of generated LFs is much lower. In our experiments, the average LF accuracy is less than 70\% in the MedAbs, ChemProt and CDR datasets, all of which require knowledge in chemical or biomedical domains. Interestingly, the average LF accuracy is much higher in the ArxivAbs dataset, although that dataset requires domain expertise in machine learning. This indicates that the accuracy of generated LFs also varies across domains. This is likely because in the case of Arxiv, papers tend to include more distinctive keywords in their abstract to match the topics for target conferences.

Next, we briefly look at other metrics. For \textit{LF\_cov\_avg} that measures the average coverage of LFs, the figures are dataset dependent. As our LFs are based on keywords or regular expressions, the LF coverage will be higher when the passages are long (e.g. IMDB) or when some indicative keywords occur frequently in the passages (e.g. Youtube). For \textit{LF\_num} that measures the total number of LFs, the figures are higher for text classification tasks than relation classification tasks, demonstrating that it is harder for PLMs to identify indicative text patterns compared to indicative keywords. The \textit{Test\_acc/F1} metric evaluates the end-to-end performance of \fname{}. While all other metrics are positively correlated with this metric, we find the \textit{PLM\_acc} $(r=0.76)$, \textit{LF\_acc\_avg} $(r=0.67)$ and \textit{Train\_acc} $(r=0.66)$ metrics have a strong positive correlation with it, evaluated using the Pearson correlation coefficient. This indicates that the accuracy of LFs is generally more important than their coverage to improve the downstream model performance.

To compare the quality of generated LFs with human-designed ones, we report the statistics for LFs in the WRENCH benchmark in Table \ref{tab:eval} as well. The average accuracy of generated LFs surpasses the LFs in WRENCH in IMDB, Yelp, Agnews and TREC datasets, showing that the evaluated PLM is good at sentiment analysis and topic classification tasks. The generated LF set usually contains more LFs, but each LF has less coverage. This is expected as many LFs in WRENCH group multiple keywords together and get activated when any of the keywords occur in the passage. With respect to the end-to-end performance, the basic few-shot prompting version of \fname{} surpasses WRENCH in YouTube, IMDB, Yelp and Spouse datasets, showing that prompting PLMs to design LFs can reach human-level performance in some tasks that require general knowledge.

In a nutshell, the evaluation indicates that the PLMs can automatically generate accurate LFs for tasks requiring general knowledge, and the generated LFs even outperform expert-designed ones in some datasets. However, the generated LFs are less accurate for tasks requiring specific domain expertise, such as chemical or biomedical knowledge. The system is also less capable of identifying text patterns to create LFs, leading to a smaller LF set for relation extraction datasets. 

\subsection{Prompting Method}
We compare the performance of \fname{} with different prompting methods: \textit{Few-Shot} for basic few-shot prompting \cite{brown2020language}, \textit{CoT} for chain-of-thought prompting \cite{wei2022chain},  and \textit{SC} for self-consistency \cite{wang2022self} on top of chain-of-thought prompting. For self-consistency, we set the number of responses as 10 and used the majority vote to get the predicted label. As each response may provide different keywords or regular expressions, we take the union of the keywords or regular expressions to create candidate LFs. Besides, we also implement the \textit{KATE} \cite{liu2021makes} method, which selects the examples that are closest to the test instance in the embedding space. We use Sentence-BERT \cite{reimers2019sentencebert} to retrieve the text embeddings and compute the cosine distances between instances. We apply \textit{KATE} together with self-consistency and report results in the \textit{SC+KATE} rows in Table \ref{tab:eval}.

We first focus on the $PLM\_acc$ metric, where the prompting methods have a direct influence. The evaluation shows that the effect of these prompting methods is highly dataset-dependent. \textit{CoT} improves the response accuracy by 10\% approximately in the TREC, ChemProt and SemEval datasets, but it leads to a similar level decrease of response accuracy in the SMS and Agnews datasets. This observation is consistent with a previous study \cite{chen2022rev} showing that chain-of-thought generates rationales that are consistent with the predicted label but do not necessarily lead to more accurate predictions. Our observation here further shows that whether chain-of-thought improves label accuracy is dataset-dependent. An interesting observation is that the decision boundaries for the SMS and Agnews datasets are blurry. For example, a promotional message may be either spam or non-spam based on whether the receiver has registered fot the company's services, while news about Google can be both related to high-tech and business. We thus hypothesize that \textit{CoT} may deteriorate prediction accuracy in datasets with fuzzy decision boundaries. One possible reason is that these datasets may occur in the PLM's training data and are memorized \cite{gehrmann2023repairing}. \textit{CoT} urges the PLM to provide labels consistent with the generated rationales, which reduces the consensus between predicted labels and labels in the training data.
\textit{SC} generally improves the PLM's response accuracy compared to \textit{CoT}, which is expected as it is essentially an ensemble of \textit{CoT}. \textit{KATE} improves the PLM's response accuracy greatly on the ChemProt and CDR datasets but reduces the PLM's response accuracy in the MedAbs dataset. This is likely because the Sentence-BERT encoder is less capable of capturing the semantic similarity in longer paragraphs.

Next, we look at the \textit{LF\_acc\_avg} metric. Generally speaking, the better the PLM performs in predicting labels, the more accurate the LFs will be. However, this positive correlation does not occur in all datasets. When we compute the correlation between $PLM\_acc$ and $LF\_acc\_avg$ separately in different datasets, the Pearson Correlation Coefficient is only positive in 6 out of 11 datasets.  One reason for that phenomenon is that wrongly annotated instances can still lead to accurate LFs. For example, the PLM may wrongly annotate "will never come again" as a positive review but design an accurate LF $come\ again\to positive$. Another reason is that the LF filters mitigate the negative effect of wrong predictions. Averaging over all datasets, no evaluated prompting method improves the LF accuracy compared to basic few-shot prompting. However, \textit{SC} consistently shows good downstream model performance. This is because different responses return different keywords or regular expressions, leading to a larger LF set with higher training data coverage. Averaging over all datasets, \textit{SC} improves the training data coverage by 10.3\% compared to \textit{Few-Shot} and improves the downstream model's performance by 3.5\% compared to \textit{Few-Shot}.

To conclude, among our evaluated prompting methods, \textit{SC} has the best overall performance in generating LFs, mainly because multiple responses help create a larger LF set with higher coverage. While \textit{CoT} and \textit{KATE} also help in some datasets, their performance is dataset-specific, and more accurate PLM responses do not always lead to more accurate LFs. Since \textit{SC} has the best overall end-to-end performance, we use this prompt for comparative studies in the following sections.

\begin{table*}[htbp]
\footnotesize
\centering
\caption{LF statistics and model performance with different pre-trained language models.} \label{tab:eval-plm}
\vspace{-0.1cm}
\resizebox{2\columnwidth}{!}{
\begin{tabular}{c c c c c c c c c c c c c c}
\toprule[1.5pt]
\textbf{Metric} & \textbf{Method} & \textbf{Youtube} & \textbf{SMS(F1)} & \textbf{IMDB} & \textbf{Yelp} & \textbf{Agnews} & \textbf{TREC} & \textbf{ArxivAbs(F1)} & \textbf{MedAbs} & \textbf{ChemProt} & \textbf{CDR(F1)} & \textbf{Spouse(F1)} & \textbf{SemEval} \\
\hline
\multirow{5}{*}{PLM\_acc} & GPT-3.5 & 83.20 (7.82) & 91.20 (3.03) & 93.20 (1.78) & 96.80 (1.09) & 86.80 (5.40) & \textbf{78.80 (6.41)} & 90.00 (3.74) & 65.60 (9.20) & 40.00 (9.48) & 63.20 (7.82) & -- & 80.40 (5.89) \\
& GPT-4 & \textbf{96.80 (3.63)} & \textbf{94.00 (8.00)} & \textbf{95.20 (1.09)} & \textbf{97.60 (1.67)} & \textbf{93.20 (2.68)} & 78.40 (6.69) & \textbf{95.60 (3.84)} & \textbf{67.60 (8.41)} & \textbf{57.20 (12.21)} & \textbf{79.20 (3.63)} & -- & \textbf{87.20 (2.28)} \\
& Llama2-CHAT-7b & 60.40 (11.34) &81.60 (7.41)& 83.20 (11.28)&80.80 (12.30)&71.20 (8.15)&26.00 (4.89)&71.20 (2.28) & 25.20 (6.72)& 4.80 (2.99)&65.60 (5.85) & -- & 10.80 (4.14) \\
& Llama2-CHAT-13b & 51.60 (9.74) & 78.00 (6.69) & 63.60 (13.22) &56.80 (5.15) &77.70 (10.98)&67.20 (7.22)& 62.40 (8.64) & 43.20 (11.19) & 8.00 (4.56)&68.40 (6.62)& -- & 54.40(4.98)
\\
& Llama2-CHAT-70b &69.60 (7.31) & 81.20 (4.48)&94.40(4.77) &96.80 (2.71)&88.80 (5.30)&62.00 (3.34) & 80.00 (6.48) & 53.60 (14.31) &19.60 (1.78)&70.00 (3.34)& -- &47.6 (28.79) \\
\hline
\multirow{5}{*}{LF\_num} & GPT-3.5 & 107.80 (9.57) & 201.00 (23.70) & 225.40 (45.60) & 246.60 (39.40) & 225.00 (20.40) & 73.60 (3.78) & 178.00 (19.82) & 87.40 (4.27) & \textbf{50.80 (8.22)}  & 28.80 (9.75) & 23.80 (17.41) & 30.40 (4.15) \\
& GPT-4 & \textbf{115.40 (7.86)} & 232.00 (29.91) & 242.40 (29.04) & 247.60 (21.26) & 256.80 (13.31) & 97.60 (4.98) & \textbf{238.40 (27.59)} & \textbf{127.00 (10.13)} & 41.00 (10.60) & 57.00 (23.30) & \textbf{38.25 (20.54)} & 36.40 (5.68) \\
& Llama2-CHAT-7b & 95.20 (18.20)& \textbf{242.80 (29.95)} & \textbf{362.60 (42.19)} & \textbf{309.00 (26.22)} &250.20 (13.06) & 31.80 (7.72) & 148.60 (36.08) & 37.80(7.79) & 4.40 (1.67)&20.20 (15.97) & 32.00(23.30) & 4.20 (1.92) \\
& Llama2-CHAT-13b & 74.40 (16.60) & 192.40 (49.02) & 215.00 (60.51) &167.40 (39.49)& \textbf{265.60 (49.22)} & \textbf{152.60 (9.31)} &186.60 (5.85) & 74.80 (6.90) & 16.80 (3.89)& \textbf{60.60 (19.98)} & 33.20 (15.18) & \textbf{42.80 (3.96)}\\
& Llama2-CHAT-70b & 98.00 (10.56) & 204.40 (19.88) &283.00(31.96) &277.00 (19.11) & 239.20 (10.53)& 126.20 (7.36)& 214.80 (34.13) & 120.60 (8.53)& 19.80 (2.49)& 9.60 (5.68) &10.60 (6.50) & 22.20(11.77)
\\
\hline
\multirow{5}{*}{LF\_acc\_avg} & GPT-3.5 & 73.50 (4.40) & 88.36 (1.03) & 72.61 (1.76) & 77.18 (1.30) & 82.41 (3.82) & \textbf{74.25 (5.73)} & 87.27 (2.05) & 61.73 (5.22) & 49.53 (3.88) & 45.26 (6.80) & -- & 71.83 (6.99) \\
& GPT-4 & \textbf{85.61 (1.25)} & \textbf{92.25 (1.09)} & \textbf{75.17 (1.12)} & \textbf{80.09 (1.78)} & \textbf{84.73 (1.89)} & 69.96 (5.41) & \textbf{89.32 (1.43)} & \textbf{63.17 (5.05)} & \textbf{53.19 (6.99)} & \textbf{69.82 (7.01)} & -- &  \textbf{80.13 (6.95)} \\
& Llama2-CHAT-7b &62.80 (6.18) & 83.32 (4.77)& 66.89 (1.87) &73.47 (0.90)& 74.37 (5.08)& 40.22 (5.04)& 80.90 (2.31)& 55.66 (9.09)&20.65 (20.18)&64.27 (5.62)& --& 65.55 (17.68) \\
& Llama2-CHAT-13b & 56.84 (9.93) & 78.39 (4.28)&67.18 (2.10) &74.99 (1.72)&78.41(4.57)&53.22 (4.40)& 77.71 (3.95) & 48.15 (7.78)&24.96 (5.81)&60.06 (2.54) & -- & 43.00 (5.82)\\
& Llama2-CHAT-70b &71.64 (5.77)& 85.25 (2.25)& 71.52 (0.96) & 76.64 (1.39)& 83.26 (2.85)&52.37 (2.55)& 84.15 (1.51) & 53.25 (7.18)&39.29 (11.10)& 47.61 (15.56) & -- & 62.56 (7.77)\\
\hline
\multirow{5}{*}{LF\_cov\_avg} & GPT-3.5 & 2.14 (0.37) & 0.86 (0.14) & \textbf{2.35 (0.40)} & \textbf{1.75 (0.25)} & 0.32 (0.03) & 0.49 (0.07) & 3.47 (0.67) & 0.62 (0.16) & 1.01 (0.10) & 1.21 (1.81) & 1.16 (1.12) & 4.57 (0.63) \\
& GPT-4 & 2.32 (0.08) & 1.07 (0.12) & 1.98 (0.25) & 1.52 (0.16) & \textbf{0.38 (0.05)} & \textbf{0.55 (0.04)} & 3.18 (0.44) & 0.84 (0.04) & \textbf{1.27 (0.21)} & 0.76 (0.68) & 1.03 (0.71) & 4.89 (0.34) \\
& Llama2-CHAT-7b & 2.10 (0.33)&\textbf{1.12 (0.20)}& 1.94 (0.18) &1.63 (0.22)&0.36(0.04)& 0.26(0.22)& \textbf{4.31 (0.77)}& \textbf{1.17 (0.25)} &0.42 (0.44)& \textbf{12.93 (12.27)} & \textbf{6.24 (6.89)} & 3.25 (1.13) \\
& Llama2-CHAT-13b &1.33 (0.25) & 0.83 (0.19) & 2.13 (0.34)& 1.52(0.13)&0.36(0.05)&0.37(0.02)& 3.30 (0.32) & 0.22 (0.09)&0.26(0.24)&3.81(1.04)& 3.12 (3.57) & 3.54 (0.26)\\
& Llama2-CHAT-70b &\textbf{2.42 (0.32)}&0.85 (0.18)& 1.95 (0.18)&1.44(0.11)&0.40(0.02)&0.42(0.03)& 3.05 (0.49) & 0.35(0.09) &1.09(0.37)&0.14(0.11) &0.47(0.32) & \textbf{5.75(3.61)} \\
\hline
\multirow{5}{*}{Train\_acc} & GPT-3.5 & 82.64 (3.05) & \textbf{92.50 (0.44)} & 79.89 (0.77) & \textbf{81.45 (1.91)} & 75.93 (9.77) & 82.65 (17.66) & 73.18 (15.78) & 58.62 (20.68) & 70.27 (9.53) & 62.79 (0.80) & -- & 83.04 (1.55) \\
& GPT-4 & \textbf{86.30 (2.22)} & 91.82 (1.31) & \textbf{80.76 (0.62)} & 80.34 (3.05) & 80.54 (0.87) & \textbf{86.92 (2.00)} & \textbf{79.12 (6.32)} & 59.03 (15.84) & \textbf{75.67 (0.34)} & \textbf{64.88 (3.39)} & -- &83.63 (1.80) \\
& Llama2-CHAT-7b & 81.98 (3.76)& 90.41 (2.63)& 78.61 (1.52)& 80.71 (1.97)& 80.69 (1.25)& 62.72 (17.09)& 66.22 (21.21)& 56.05 (20.98) &32.71 (29.17)&59.43 (7.29) & -- & \textbf{91.61 (2.94)} \\
& Llama2-CHAT-13b &77.13 (10.91) & 87.74 (0.42) & 75.55 (0.35)& 80.75 (0.81)&72.08 (17.78)&78.66(3.65)& 57.80 (24.39) & 54.51 (17.17) &50.45 (20.59)&57.47 (4.57) & -- & 79.80 (1.58)\\
& Llama2-CHAT-70b &82.65 (1.39)& 91.92 (1.22)&78.60 (2.57) &78.75 (4.02)& \textbf{81.76 (0.63)}&68.18 (27.92) & 65.89 (11.82)& \textbf{67.27 (1.26)} & 67.32 (8.00)&55.90 (3.68) & --&80.82 (0.81)\\
\hline
\multirow{5}{*}{Train\_cov} & GPT-3.5 & 81.97 (1.48) & 75.38 (6.03) & 97.73 (0.60) & 94.65 (1.58) & 47.99 (4.42) & 29.84 (2.52) & 99.69 (0.20) & 36.45 (5.74) & \textbf{31.00 (3.08)} & 100.00 (0.00) & 100.00 (0.00) & \textbf{77.78 (7.28)} \\
& GPT-4 & \textbf{83.78 (1.28)} & 85.95 (7.27) & 97.19 (0.61) & 92.98 (1.11) & \textbf{59.04 (4.10)} &  40.08 (3.42) & \textbf{99.89 (0.04)} & \textbf{49.98 (1.81)} & 30.21 (1.92) & 100.00(0.00) & 100.00 (0.00) & 77.68 (3.15) \\
& Llama2-CHAT-7b & 74.34 (6.78)& \textbf{86.86 (7.05)}& \textbf{98.85 (0.53)} &\textbf{95.56 (1.68)}& 55.96 (3.00)& 7.92 (7.66)& 99.8 (0.10)& 23.9 (1.64)&2.30 (2.63)& 100.00 (0.00) & 100.00 (0.00) & 7.63 (3.03) \\
& Llama2-CHAT-13b & 57.82 (13.82)& 72.50 (14.89)& 95.15 (2.92)&85.54 (6.60)&56.97 (9.04)&\textbf{44.52 (2.92)}& 99.76 (0.17)& 13.45 (5.19)& 3.50(3.22)&100.00 (0.00) & 100.00 (0.00) &76.50 (4.85) \\
& Llama2-CHAT-70b & 78.83 (4.31)& 76.12 (10.15) &98.03 (0.28) &93.48 (1.70) & 57.57 (2.00)&42.49 (2.60)& 99.44 (0.94) & 30.00 (5.61) &20.48 (7.07)&100.00 (0.00) & 100.00 (0.00) &66.34 (15.24)\\
\hline
\multirow{5}{*}{Test\_acc/F1} & GPT-3.5 & 87.92 (3.48) & 82.88 (2.89) & 79.39 (0.41) & \textbf{87.39 (0.67)} & 75.84 (11.82) & 45.96 (8.57) & 57.95 (15.59) & \textbf{39.72 (0.85)} & 41.84 (6.44) & 21.65 (11.61) & 36.25 (9.08) & \textbf{81.10 (5.27)} \\
& GPT-4 & \textbf{89.84 (3.43)} & 85.26 (1.48) & \textbf{80.06 (0.23)} & 87.10 (1.02) & \textbf{81.85 (1.30)} & 56.52 (2.54) & \textbf{64.00 (11.19)} & 39.52 (1.71) & \textbf{43.75 (1.96)} & \textbf{33.70 (12.07)} & 34.24 (9.38) & 74.53 (0.44) \\
& Llama2-CHAT-7b & 89.52 (3.39)&81.25 (6.09)& 78.58 (0.56) &86.90 (0.66)&81.02 (1.14)& 36.76 (5.84)& 49.95 (27.61) & 27.30 (4.73)&21.44 (8.93)&24.1 (14.71) & 7.36 (14.73) & 17.17 (0.00) \\
& Llama2-CHAT-13b &82.60 (9.45)&80.76 (5.39)& 77.54 (1.34) &86.81 (0.72) &71.93 (20.98)& \textbf{57.88 (5.26)} &43.29 (21.74) & 35.10 (3.03) &18.15 (8.18)& 24.32 (15.18) & \textbf{36.40 (5.65)} &71.83 (4.93)\\
& Llama2-CHAT-70b & 87.12 (2.16)&\textbf{85.53 (2.49)}&78.55 (0.88) &86.44 (1.67) & 81.71 (1.05)&50.20 (15.61)&54.49 (4.48) & 39.55 (1.12)&33.04 (2.57)&13.42 (13.70) & 23.89 (22.02) & 64.77 (13.18)\\
\bottomrule[1.5pt]
\end{tabular}}
\end{table*}

\subsection{PLM Model}

Next, we evaluate different pre-trained language models in \fname{}. For GPT-3.5 and GPT-4, we use the \textit{gpt-3.5-turbo-0613} and \textit{gpt-4-0613} models from the OpenAI API \cite{openai2023}. For Llama2-CHAT models \footnote{Llama2-CHAT models are Llama2 models fine-tuned for dialogue use cases.}, we use the API of Anyscale AI\cite{anyscale2023} and evaluate three variants of Llama2-CHAT models with 7 billion, 13 billion and 70 billion parameters, respectively. The results are illustrated in table \ref{tab:eval-plm}. 

Looking at the downstream model's performance, the GPT-4 model has the best performance on average and surpasses other models in 6 out of 12 evaluated datasets. The Llama2-CHAT models' end-to-end performance is comparable to GPT-3.5, especially on simple tasks such as sentiment analysis and topic classification. For domain-specific datasets, all models have relatively poor performance. However, when we look at the PLM's response accuracy, the Llama2-CHAT-70b model is comparable to GPT-3.5, while the Llama2-CHAT-7b and Llama2-CHAT-13b models underperform other models with a large gap. After checking their responses, we find that the Llama2-CHAT-7b  and Llama2-CHAT-13b models have more problems following the instruction format. They frequently create artificial examples and design LFs based on them in the responses instead of answering the query in the user prompt. The Llama2-CHAT-70b model also generates artificial examples sometimes, but it usually responds to the user query first. This observation is consistent with previous work \cite{liu2023agentbench}, which shows open-sourced models like Llama2 make more mistakes in following instruction formats compared to commercial API-based models. 

In summary, our evaluation shows that the Llama2-CHAT-70b model is a good substitute for more expensive commercial models in simple tasks, while the smaller Llama2-CHAT-7b and Llama2-CHAT-13b models have problems following the instruction format. For tasks requiring domain expertise, all evaluated models perform poorly. This indicates that exploring the applicability of specialized LLMs \cite{luo2022biogpt,wu2023bloomberggpt} to design LFs for domain-specific datasets is a promising research direction. 

\subsection{Instance Selection}
\begin{table*}[htbp]
\footnotesize
\centering
\caption{LF statistics and model performance with different instance selection methods.} \label{tab:eval-sample}
\vspace{-0.1cm}
\resizebox{2\columnwidth}{!}{
\begin{tabular}{c c c c c c c c c c c c c c}
\toprule[1.5pt]
\textbf{Metric} & \textbf{Method} & \textbf{Youtube} & \textbf{SMS(F1)} & \textbf{IMDB} & \textbf{Yelp} & \textbf{Agnews} & \textbf{TREC} & \textbf{ArxivAbs(F1)} & \textbf{MedAbs} & \textbf{ChemProt} & \textbf{CDR(F1)} & \textbf{Spouse(F1)} & \textbf{SemEval} \\
\hline
\multirow{3}{*}{PLM\_acc} & random &  \textbf{83.20 (7.82)} & 91.20 (3.03) & 93.20 (1.78) & \textbf{96.80 (1.09)} & \textbf{86.80 (5.40)} & 78.80 (6.41) & 90.00 (3.74) & \textbf{65.60 (9.20)} & \textbf{40.00 (9.48)} & 63.20 (7.82) & -- & 80.40 (5.89) \\
& uncertain & 80.40 (3.84) & 84.40 (5.36) & 91.20 (4.38) & 94.40 (3.28) & 73.60 (4.56) & 74.40 (4.33) & 89.60 (2.60) & 62.80 (5.76) & 34.80 (6.09) & \textbf{67.20 (9.75)} & -- & 80.80 (6.41) \\
& SEU & 58.80 (49.17) & \textbf{92.00 (7.07)} & \textbf{94.40 (0.89)} & 85.60 (3.84) & 68.40 (6.54) & \textbf{90.80 (3.03)} & \textbf{94.00 (2.00)} & 56.80 (1.78) & 37.20 (3.63) & 61.60 (5.17) & -- & \textbf{89.20 (2.68)} \\
\hline
\multirow{3}{*}{LF\_num} & random &  107.80 (9.57) & \textbf{201.00 (23.70)} & \textbf{225.40 (45.60)} & \textbf{246.60 (39.40)} & \textbf{225.00 (20.40)} & \textbf{73.60 (3.78)} & \textbf{178.00 (19.82)} & \textbf{87.40 (4.27)} & \textbf{50.80 (8.22)} & \textbf{28.80 (9.75)} & 23.80 (17.41) & 30.40 (4.15) \\
& uncertain & \textbf{118.00 (2.91)} & 191.80 (11.25) &  224.80 (43.40) & 232.40 (22.36) & 193.20 (16.17) & 73.00 (10.27) & 163.60 (17.89) & 82.40 (9.99) & 49.20 (11.27) & 28.20 (8.28) & \textbf{46.20 (34.70)} & \textbf{41.80 (4.65)} \\
& SEU & 24.80 (19.61) & 62.00 (8.77) & 112.00 (15.37) & 77.20 (17.81) & 125.60 (9.76) & 3.40 (0.89) & 130.00 (6.48) & 64.40 (2.30) & 35.40 (5.32) & 15.80 (6.09) & 16.20 (9.96) & 13.60 (2.70) \\
\hline
\multirow{3}{*}{LF\_acc\_avg} & random & 73.50 (4.40) & 88.36 (1.03) & 72.61 (1.76) & 77.18 (1.30) & \textbf{82.41 (3.82)} & \textbf{74.25 (5.73)} & 87.27 (2.05) & 61.73 (5.22) & 49.53 (3.88) & 45.26 (6.80) & -- & \textbf{71.83 (6.99)} \\
& uncertain & 69.31 (2.82) & 81.09 (6.81)& 72.74 (1.78) & \textbf{77.62 (1.00)} & 73.76 (0.68) & 67.39 (7.56) & \textbf{87.50 (1.83)} & \textbf{63.58 (3.91)} & \textbf{50.26 (4.45)} & 53.37 (3.90) & -- & 58.17 (2.76) \\
& SEU & \textbf{78.77 (5.45)} & \textbf{94.44 (0.75)} & \textbf{73.25 (0.93)} & 77.23 (2.57) & 75.40 (1.24) & 65.57 (21.53) & 86.94 (0.92) & 58.22 (1.20) & 49.20 (4.36) & \textbf{53.83 (9.82)} & -- & 70.55 (7.14) \\
\hline
\multirow{3}{*}{LF\_cov\_avg} & random & 2.14 (0.37) & 0.86 (0.14) & 2.35 (0.40) & 1.75 (0.25) & \textbf{0.32 (0.03)} & \textbf{0.49 (0.07)} & 3.47 (0.67) & \textbf{0.62 (0.16)} & 1.01 (0.10) & 1.21 (1.81) & 1.16 (1.12) & 4.57 (0.63) \\
& uncertain & 1.83 (0.15) & 0.76 (0.11) & 2.15 (0.42) & 1.82 (0.35) &  0.24 (0.05) & 0.41 (0.06) & 3.80 (0.69) & 0.45 (0.18) & 0.91 (0.07) & 0.83 (1.34) & \textbf{2.55 (3.91)} & 2.94 (0.60) \\
& SEU & \textbf{3.40 (1.84)} & \textbf{1.81 (0.17)} & \textbf{3.28 (0.39)} & \textbf{2.53 (0.67)} & \textbf{0.37 (0.03)} & 0.26 (0.14) & \textbf{4.50 (0.41)} & 0.50 (0.17) & \textbf{1.17 (0.16)} & \textbf{1.81 (2.60)} & 1.20 (1.19) & \textbf{6.54 (1.55)} \\
\hline
\multirow{3}{*}{Train\_acc} & random & 82.64 (3.05) & \textbf{92.50 (0.44)} & \textbf{79.89 (0.77)} & \textbf{81.45 (1.91)} & 75.93 (9.77) & 82.65 (17.66) & \textbf{73.18 (15.78)} & 58.62 (20.68) & 70.27 (9.53) & 62.79 (0.80) & -- & \textbf{83.04 (1.55)} \\
& uncertain & \textbf{85.40 (1.45)} & 88.55 (4.76) & 79.23 (0.83) & 81.35 (0.52) & 70.46 (15.89) & \textbf{88.22 (3.35)} & 63.16 (24.03) & 67.32 (2.13) & 73.29 (2.83) & \textbf{63.92 (2.45)} & -- & 75.00 (0.95) \\
& SEU & 80.42 (5.86) & 89.08 (1.20) & 78.33 (0.67) & 70.32 (5.62) & \textbf{77.94 (1.28)} & 63.83 (13.72) & 76.42 (3.27) & \textbf{68.86 (1.12)} & \textbf{75.13 (0.25)} & 63.51 (5.25) & -- & 79.05 (3.32) \\
\hline
\multirow{3}{*}{Train\_cov} & random & 81.97 (1.48) & \textbf{75.38 (6.03)} & \textbf{97.73 (0.60)} & \textbf{94.65 (1.58)} & \textbf{47.99 (4.42)} &\textbf{ 29.84 (2.52)} & \textbf{99.69 (0.20)} & \textbf{36.45 (5.74)} & \textbf{31.00 (3.08)} & 100.0 (0.00) & 100.0 (0.00) & \textbf{77.78 (7.28)} \\
& uncertain & \textbf{83.24 (1.36)} & 69.24 (4.44) & 97.09 (0.38) & 94.21 (2.20) & 35.81 (7.59) & 25.81 (2.78) & 99.55 (0.30) & 27.81 (7.12) & 30.68 (3.38) & 100.0 (0.00) & 100.0 (0.00) & 68.40 (9.98) \\
& SEU & 50.10 (36.27) & 65.18 (5.79) & 94.62 (0.86) & 79.87 (5.55) & 36.24 (1.86) & 0.94 (0.67) & 99.51 (0.12) & 26.14 (8.30) & 28.76 (1.20) & 100.0 (0.00) & 100.0 (0.00) & 51.94 (19.14) \\
\hline
\multirow{3}{*}{Test\_acc/F1} & random & \textbf{87.92 (3.48)} & \textbf{82.88 (2.89)} & \textbf{79.39 (0.41)} & 87.39 (0.67) & 75.84 (11.82) & 45.96 (8.57) & 57.95 (15.59) & \textbf{39.72 (0.85)} & 41.84 (6.44) & 21.65 (11.61) & \textbf{36.25 (9.08)} & \textbf{81.10 (5.27)} \\
& uncertain & 87.76 (2.70) & 82.34 (3.55) & 79.30 (0.71) & \textbf{87.65 (0.58)} & 73.85 (15.41) & \textbf{53.24 (11.28)} & 49.35 (19.48) & 37.89 (1.14) & \textbf{47.34 (4.08)} & 23.80 (18.12) & 34.10 (22.94) & 75.00 (0.95) \\
& SEU & 81.00 (16.69) & 72.96 (15.70) & 78.69 (0.34) & 83.54 (2.51) & \textbf{78.63 (2.32)} & 18.76 (2.90) & \textbf{63.04 (7.09)} & 36.39 (1.16) & 41.16 (2.75) & \textbf{32.18 (10.94)} & 31.69 (17.99) & 59.47 (15.97)\\
\bottomrule[1.5pt]
\end{tabular}}
\end{table*}

Table \ref{tab:eval-sample} evaluates the influence of query instance selection methods on the performance of \fname{}. While the best selection method is dataset-dependent, we find that random sampling generally performs well in our experiments, which leads to the best downstream model performance in 6 out of 12 evaluated datasets. As random sampling selects a diverse set of instances for PLMs to provide label functions, it helps \fname{} to get a larger LF set and increases the training data coverage, which improves the performance of the downstream model.

While uncertain sampling is a popular method for active learning, it generally underperforms random sampling in our evaluation. One reason for that is that traditional active learning assumes a perfect oracle, while our framework leverages PLMs that are imperfect. Uncertain sampling selects instances where the current model is most uncertain, and these instances tend to be more difficult than average for labelling. In 9 out of 12 evaluated datasets, the PLM's response accuracy using uncertainty sampling reduces compared to using random sampling. This leads to reduced LF accuracy and hurts downstream model performance. The SEU sampler, on the other hand, assumes one would return LFs with probabilities proportional to LF accuracy, which may not fit the behaviour of the PLM applied in \fname{}. For example, in the TREC dataset, SEU generates much fewer LFs than the other two samplers. We find that SEU selects questions that start with "what is" repeatedly, indicating that its user model believes that phrase is a helpful indicator to build LFs. However, when \fname{} does not generate LFs based on that phrase, SEU cannot learn from the feedback but select similar instances again, leading to performance degradation. 

Our evaluation indicates that while query instance selection has a significant influence on the performance of \fname{}, current selection methods do not work well for LF development with PLMs as they do not consider the imperfect nature of PLMs or fail to learn from PLM feedbacks. Further research is required to develop instance selection methods that take these factors into consideration. 

\subsection{LF filtering}
\begin{table*}[htbp]
\footnotesize
\centering
\caption{LF statistics and model performance with different LF filtering methods.} \label{tab:eval-filter}
\vspace{-0.1cm}
\resizebox{2\columnwidth}{!}{
\begin{tabular}{c c c c c c c c c c c c c c}
\toprule[1.5pt]
\textbf{Metric} & \textbf{Method} & \textbf{Youtube} & \textbf{SMS(F1)} & \textbf{IMDB} & \textbf{Yelp} & \textbf{Agnews} & \textbf{TREC} & \textbf{ArxivAbs(F1)} & \textbf{MedAbs} & \textbf{ChemProt} & \textbf{CDR(F1)} & \textbf{Spouse(F1)} & \textbf{SemEval} \\
\hline
\multirow{3}{*}{PLM\_acc} & all &  \textbf{83.20 (7.82)} & \textbf{91.20 (3.03)} & 93.20 (1.78) & \textbf{96.80 (1.09)} & 86.80 (5.40) & \textbf{78.80 (6.41)} & 90.00 (3.74) & 65.60 (9.20) & \textbf{40.00 (9.48)} & 63.20 (7.82) & -- & \textbf{80.40 (5.89)} \\
& no accuracy &  \textbf{83.20 (5.93)} & 88.40 (2.60) & 92.40 (2.96) & \textbf{96.80 (1.09)} & \textbf{87.20 (3.63)} & 74.80 (4.60) & 90.40 (4.98) & 57.60 (7.92) & 36.40 (4.56) & 66.80 (6.26) & -- & 79.20 (1.78) \\
& no redundancy & 82.80 (8.31) & 89.60 (3.28) & \textbf{94.40 (1.67)} & \textbf{96.80 (1.09)} & \textbf{87.20 (3.63)} & 76.40 (8.87) & \textbf{91.60 (3.84)} & \textbf{67.20 (9.23)} & 39.20 (8.67) & \textbf{69.60 (9.52)} & -- & 78.80 (5.76) \\
\hline
\multirow{3}{*}{LF\_num} & all &  107.80 (9.57) & 201.00 (23.70) & 225.40 (45.60) & 246.60 (39.40) & 225.00 (20.40) & 73.60 (3.78) & 178.00 (19.82) & 87.40 (4.27) & 50.80 (8.22) & 28.80 (9.75) & 23.80 (17.41) & 30.40 (4.15) \\
& no accuracy & 128.80 (8.87) & 234.00 (20.84) & \textbf{368.20 (60.83)} & \textbf{371.00 (67.14)} & \textbf{324.80 (10.25)} & 96.00 (12.08) & 220.20 (29.55) & \textbf{182.00 (8.21)} & \textbf{198.00 (15.76)} & \textbf{46.80 (14.36)} & 36.40 (17.27) & \textbf{68.00 (7.74)} \\
& no redundancy & \textbf{209.60 (34.91)} & \textbf{259.20 (29.15)} & 294.00 (59.98) & 337.40 (46.42) & 260.80 (29.98) & \textbf{113.80 (7.08)} &  \textbf{241.20 (34.47)} & 108.80 (9.01) & 76.40 (9.42) & 39.80 (16.54) & 37.40 (24.58) & 58.00 (5.52) \\
\hline
\multirow{3}{*}{LF\_acc\_avg} & all & 73.50 (4.40) & 88.36 (1.03) & 72.61 (1.76) & 77.18 (1.30) & \textbf{82.41 (3.82)} & 74.25 (5.73) & 87.27 (2.05) & 61.73 (5.22) & 49.53 (3.88) & 45.26 (6.80) & -- & \textbf{71.83 (6.99)} \\
& no accuracy & 69.00 (1.36) & 77.29 (2.87) & 63.97 (2.01) & 68.51 (1.08) & 67.79 (4.15) & 60.14 (5.14) & 76.14 (3.34) & 46.37 (4.36) & 20.16 (1.96) & \textbf{46.87 (7.70)} & -- & 31.47 (2.28) \\
& no redundancy & \textbf{82.20 (3.32)} & \textbf{88.44 (1.10)} & \textbf{73.20 (1.45)} & \textbf{77.63 (0.68)} & 82.20 (4.94) & \textbf{75.99 (7.15)} & \textbf{88.06 (1.62)} & \textbf{62.70 (7.30)} & \textbf{59.28 (5.33)} & 41.87 (12.57) & -- & 69.45 (5.90) \\
\hline
\multirow{3}{*}{LF\_cov\_avg} & all & 2.14 (0.37) & 0.86 (0.14) & 2.35 (0.40) & 1.75 (0.25) & 0.32 (0.03) & 0.49 (0.07) & 3.47 (0.67) & 0.62 (0.16) & 1.01 (0.10) & 1.21 (1.81) & 1.16 (1.12) & 4.57 (0.63) \\
& no accuracy & 2.64 (0.33) & \textbf{1.10 (0.24)} & 3.29 (0.30) & 2.38 (0.19) & \textbf{0.48 (0.04)} &  \textbf{2.69 (0.76)} & 4.18 (0.43) & \textbf{1.01 (0.20)} & \textbf{2.28 (0.69)} & \textbf{2.44 (1.25)} & \textbf{1.59 (1.86)} & \textbf{10.87 (1.34)} \\
& no redundancy & \textbf{4.52 (1.49)} & 0.83 (0.06) & \textbf{4.21 (0.54)} & \textbf{3.13 (0.27)} & 0.37 (0.03) & 0.63 (0.19) & \textbf{4.62 (0.72)} & 0.70 (0.27) & 2.10 (0.52) & 1.11 (1.82) & 1.58 (2.78) & 8.22 (1.64) \\
\hline
\multirow{3}{*}{Train\_acc} & all & \textbf{82.64 (3.05)} & \textbf{92.50 (0.44)} & \textbf{79.89 (0.77)} & \textbf{81.45 (1.91)} & 75.93 (9.77) & 82.65 (17.66) & \textbf{73.18 (15.78)} & \textbf{58.62 (20.68)} & 70.27 (9.53) & 62.79 (0.80) & -- & \textbf{83.04 (1.55)} \\
& no accuracy & 69.76 (12.17) & 86.20 (6.75) & 65.46 (12.26) & 73.11 (5.35) & 61.88 (5.22) & 42.35 (2.95) &  54.08 (27.90) & 45.32 (8.97) & 35.09 (5.35) & 55.30 (6.21) & -- & 43.86 (23.03) \\
& no redundancy & 82.12 (3.82) & 91.91 (1.00) & 75.88 (2.02) & 76.07 (2.76) & \textbf{79.76 (1.23)} & \textbf{90.92 (1.90)} & 71.04 (15.69) & 58.28 (20.69) & \textbf{71.25 (8.73)} & \textbf{63.71 (1.96)} & -- & 74.43 (8.94) \\
\hline
\multirow{3}{*}{Train\_cov} & all & 81.97 (1.48) & 75.38 (6.03) & 97.73 (0.60) & 94.65 (1.58) & 47.99 (4.42) & 29.84 (2.52) & 99.69 (0.20) & 36.45 (5.74) & 31.00 (3.08) & 100.0 (0.00) & 100.0 (0.00) & 77.78 (7.28) \\
& no accuracy & \textbf{86.37 (1.51)} &\textbf{ 80.88 (4.99)} & \textbf{99.83 (0.13)} & \textbf{98.39 (0.41)} &  \textbf{74.30 (3.19)} & \textbf{88.61 (2.21)} & \textbf{99.93 (0.03)} & \textbf{76.44 (7.55)} & \textbf{82.67 (18.24)} & 100.0 (0.00) & 100.0 (0.00) & \textbf{99.06 (0.74)} \\
& no redundancy & 82.27 (1.04) & 75.62 (5.41) & 97.42 (0.51) & 94.16 (1.29) & 48.17 (3.69) & 29.43 (2.36) & 99.75 (0.15) & 35.78 (5.67) & 30.22 (3.00) & 100.0(0.00) & 100.0(0.00) & 72.11 (7.50) \\
\hline
\multirow{3}{*}{Test\_acc/F1} & all & \textbf{87.92 (3.48)} & 82.88 (2.89) & 79.39 (0.41) & \textbf{87.39 (0.67)} & 75.84 (11.82) & 45.96 (8.57) & 57.95 (15.59) & 39.72 (0.85) & 41.84 (6.44) & 21.65 (11.61) & 36.25 (9.08) & \textbf{81.10 (5.27)} \\
& no accuracy & 80.16 (10.07) & 74.38 (9.10) & 69.11 (13.61) & 83.39 (2.30) & 74.80 (6.00) & 35.28 (8.66) & 58.75 (15.57) & 39.20 (4.32) & 34.39 (3.48) & \textbf{38.54 (8.54)} & 34.58 (4.77) & 49.57 (25.80) \\
& no redundancy & 85.12 (2.65) & \textbf{83.24 (2.73)} & \textbf{79.57 (0.86)} & 85.52 (1.18) & \textbf{80.17 (2.08)} & \textbf{55.04 (10.11)} &  \textbf{59.23 (19.86)} & \textbf{39.85 (0.92)} & \textbf{45.19 (0.74)} & 24.31(14.26) & \textbf{42.42 (8.28)} & 67.80 (6.56)\\
\bottomrule[1.5pt]
\end{tabular}}
\end{table*}

In this section, we evaluate the influence of LF filtering on the performance of \fname{}. We compare three versions of LF filtering: apply all filters, ablate the accuracy filter and ablate the redundancy filter. Table \ref{tab:eval-filter} shows the results. Ablating either the accuracy filter or the redundancy filter increases the total number of LFs and improves the total coverage of the LF set. Ablating the accuracy filter leads to a significant decrease in average LF accuracy in most evaluated datasets and results in lower downstream model performance. On the contrary, the ablation of the redundancy filter increases the average LF accuracy and improves the downstream model performance in most evaluated datasets. This is likely because some high-quality LFs will be generated repeatedly by the PLM.  However, training a label model on an LF set with high redundancy can make the model more complicated and difficult to interpret. Our evaluation suggests that the accuracy filter should always be applied to guarantee downstream model performance. Whether the application of the redundancy filter will be helpful or not is dataset-dependent.

\section{Discussions}
Based on our evaluation, we provide a few takeaways for leveraging PLMs to design LFs and point out several future research directions.

\begin{description}
\item[T1:] PLMs can achieve good performance in designing keyword-based LFs for tasks requiring general knowledge. However, they are less effective in designing pattern-based LFs, and they are also less accurate in tasks requiring specific domain expertise. \textbf{(RQ1)}

\item[T2:] Overall, GPT-4 performs the best for designing LFs, followed by GPT-3.5 and Llama2-CHAT-70b models. This makes Llama2-CHAT-70b a cost effective alternative to commercial models. However, Llama2-CHAT-7b and Llama2-CHAT-13b models have problems following the formats specified in the prompts. \textbf{(RQ1)}

\item[T3:] While prompting methods like chain-of-thought and \textit{KATE} improve the PLM's accuracy in some datasets, their performance is dataset-dependent, and more accurate predictions do not always lead to more accurate LFs. However, asking the PLM to provide multiple responses generally helps to create more candidate LFs and improve the end-to-end system performance. \textbf{(RQ2)}

\item[T4:] Current active query instance selection methods do not work well in prompting PLMs to design LFs, because they do not consider the imperfect nature of PLMs or learn from PLM feedback. Further research is required to design more effective selection methods for PLM prompting. \textbf{(RQ3)}

\item[T5:] The variance across different runs is large in some evaluated datasets. This is because many components in the framework are non-deterministic, such as the selection of query instances, the selection of in-context examples, and the PLM's responses. Future research can explore improvements to the robustness of the framework by reducing variance in these steps. \textbf{(RQ4)}

\item[T6:] In some datasets, the number of LFs for some classes will be much higher than LFs for other classes. We refer to this as \textit{the LF class imbalance issue}. There are two possible sources for it: either the training dataset has a skewed label distribution, or it is easier for the PLM to generate LFs for some classes than others. In this work, we apply the \textit{default-class} technique to mitigate the issue. Future research can focus on addressing the LF class imbalance issue further. \textbf{(RQ4)}

\end{description}

\section{Conclusions}
In this paper, we investigated the potential of PLMs in designing accurate LFs for the data programming framework. We designed an interactive framework \fname{} that can automatically develop LFs by prompting PLMs, and extensively evaluated \fname{} in 12 real-world datasets. We explore the design space thoroughly, identifying the strengths and limitations of PLMs in designing LFs. Finally, we point out several promising research directions for future works.

\bibliographystyle{ACM-Reference-Format}
\bibliography{reference}

\end{document}